\pdfoutput=1

\documentclass[11pt]{article}

\usepackage[final]{acl}

\usepackage{times}
\usepackage{latexsym}

\usepackage[T1]{fontenc}

\usepackage[utf8]{inputenc}

\usepackage{microtype}

\usepackage{inconsolata}
\usepackage{adjustbox}
\usepackage{booktabs}
\usepackage{arydshln}
\usepackage{makecell}
\usepackage{float}

\newcommand{\dataset}{MDCR}
\newcommand{\croucher}{\textsl{the Croucher Studentship}}
\newcommand{\ms}{\textsl{the Microsoft Scholarship}}
\newcommand{\coca}{\textsl{the Coca-Cola Scholarship}}
\newcommand{\circlednum}[1]{\raisebox{.5pt}{\textcircled{\raisebox{-.9pt} {#1}}}}

\title{\dataset{}: A Dataset for Multi-Document Conditional Reasoning}

\author{
Peter Baile Chen, Yi Zhang*, Chunwei Liu, Sejal Gupta, Yoon Kim, Michael Cafarella \\
MIT \qquad *AWS AI Labs \\
\{peterbc, chunwei, sejalg, yoonkim\}@mit.edu, imyi@amazon.com, michjc@csail.mit.edu
}

\begin{document}
\maketitle
\begin{abstract}
The same real-life questions posed to different individuals may lead to different answers based on their unique situations. For instance, whether a student is eligible for a scholarship depends on eligibility conditions, such as major or degree required.
ConditionalQA was proposed to evaluate models' capability of 
reading a document and answering eligibility questions, considering \textit{unmentioned} conditions.
However, it is limited to questions on single documents, neglecting harder cases that may require \emph{cross-document reasoning} and \emph{optimization}, for example, ``What is the maximum number of scholarships attainable?'' Such questions over multiple documents are not only more challenging due to more context having to understand, but also because the model has to (1) explore all possible combinations of unmentioned conditions and (2) understand the relationship between conditions across documents, to reason about the optimal outcome.
To evaluate models' capability of answering such questions, we propose a new dataset \dataset{}, which 
can reflect real-world challenges and serve as a new test bed for complex conditional reasoning that requires optimization. We evaluate this dataset using the most recent LLMs and demonstrate their limitations in solving this task. We believe this dataset will facilitate future research in answering optimization questions with unknown conditions. \footnote{Datasets and code will be released upon acceptance.}
\end{abstract}

\section{Introduction}\label{sec:intro}



\begin{figure}
\centering
\includegraphics[width=0.54\textwidth]{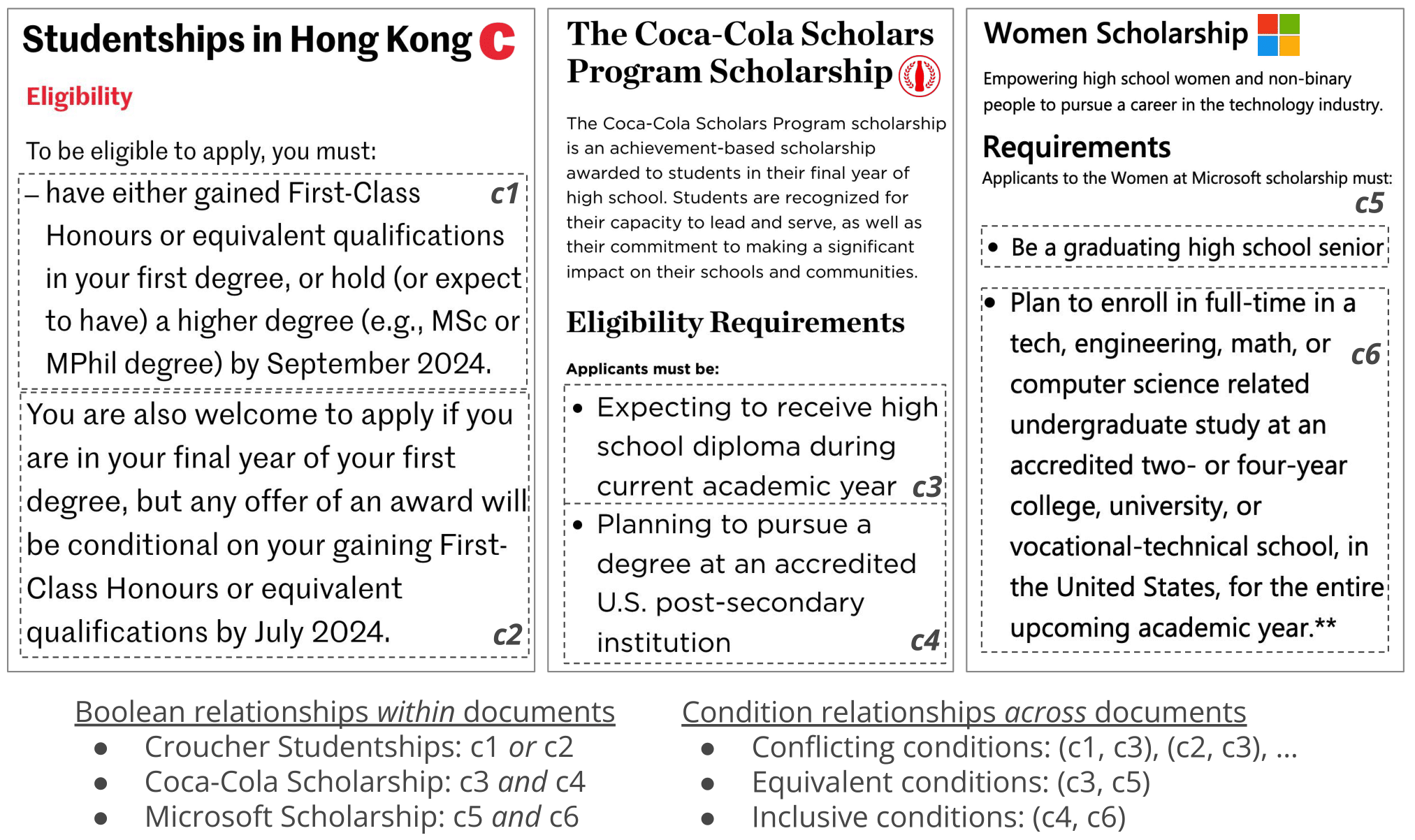}
\caption{An example of documents and relationships.}
\label{fig:diagram1}
\end{figure}

Answers to many real-life questions depend on geographical or temporal situations~\cite{min2020ambigqa,zhang2021situatedqa,stelmakh2023asqa} or the facts about the person who asked the question~\cite{sun-etal-2022-conditionalqa}. Consider a high school senior student reading a scholarship document that describes its eligibility conditions. 
The answer to whether the student is eligible for the program depends on, for instance, whether the student is ``planning to pursue a degree in a US post-secondary education''. The answer is yes if this condition is satisfied and no otherwise. This condition, despite being \textit{unmentioned} in the question, is necessary for the ``yes'' answer to hold.

Scholarships, internships, and government benefits, among others, are popular domains where (eligibility) conditions frequently appear in documents and eligibility questions naturally occur.
ConditionalQA \cite{sun-etal-2022-conditionalqa} was proposed to examine the performance of models for such questions on single benefit documents in the public policy domain. 
Yet, besides binary yes/no questions on single documents, users are also interested in asking maximization questions over multiple benefits that require \textit{optimization}.
For instance, students can ask questions about application strategies to maximize the number of scholarships attainable to cover tuition.
Low-income families can ask about qualifications to maximize their social benefits (e.g., tax credit, housing allowances) to improve living standards.
Existing datasets that target questions on single documents neglect these common questions over \emph{multiple} documents~\cite{kulkarni2020aquamuse,Boni2021HowSummAM,bolotova-baranova-etal-2023-wikihowqa}.
These questions that need to consider \textit{unmentioned} conditions over \textit{multiple documents jointly} 
 pose several new challenges.

First, it requires a fine-grained multi-document understanding. In particular, models not only need to understand the conditions in each document, but also the \emph{relationships} between conditions across documents, which are critical to answering the multi-document questions correctly. Figure \ref{fig:diagram1} shows three common relationships: \emph{conflicting}, \emph{equivalent}, and \emph{inclusive}.
\begin{itemize}
\item
\textit{Conflicting}: \href{https://croucher.org.hk/en/funding/study_awards/hk-studentships}{\croucher{}} require applicants to \textit{``have either gained first-class honors ... or hold ... a higher degree ...''}, which are qualifications attained in post-secondary education. These conditions conflict with the condition \textit{``expecting to receive high school diploma''} (a qualification attained in secondary education) as required in the \href{https://www.coca-colascholarsfoundation.org/apply/}{\coca{}}. Therefore, a user \textit{cannot be} eligible for both scholarships. Models can only make such conclusions if they can compare conditions across documents.
\item 
\textit{Equivalent}:
\coca{}'s condition \textit{``expecting to receive a high school diploma ...''} is equivalent to the condition \textit{``be a graduating high school senior''} as stated in \href{https://www.microsoft.com/en-us/diversity/programs/women-at-microsoft-scholarship}{\ms{}}.
Satisfying either naturally translates to satisfying both.
\item 
\textit{Inclusive}:
The condition \textit{``Planning to pursue a degree at an accredited U.S. post-secondary institution''} in \coca{} is inclusive of the condition \textit{``Plan to enroll in full-time in a [STEM] related undergraduate study at a ... college ... in the United States ...''} in \ms{} because both require applicants to enroll in post-secondary education, but the latter is more \textit{restrictive} in terms of major and degree.
Therefore, satisfying the latter means the former is satisfied; if the former is unsatisfied, the latter is also unsatisfied.
\end{itemize}

\begin{figure}
\centering
\includegraphics[width=\columnwidth]{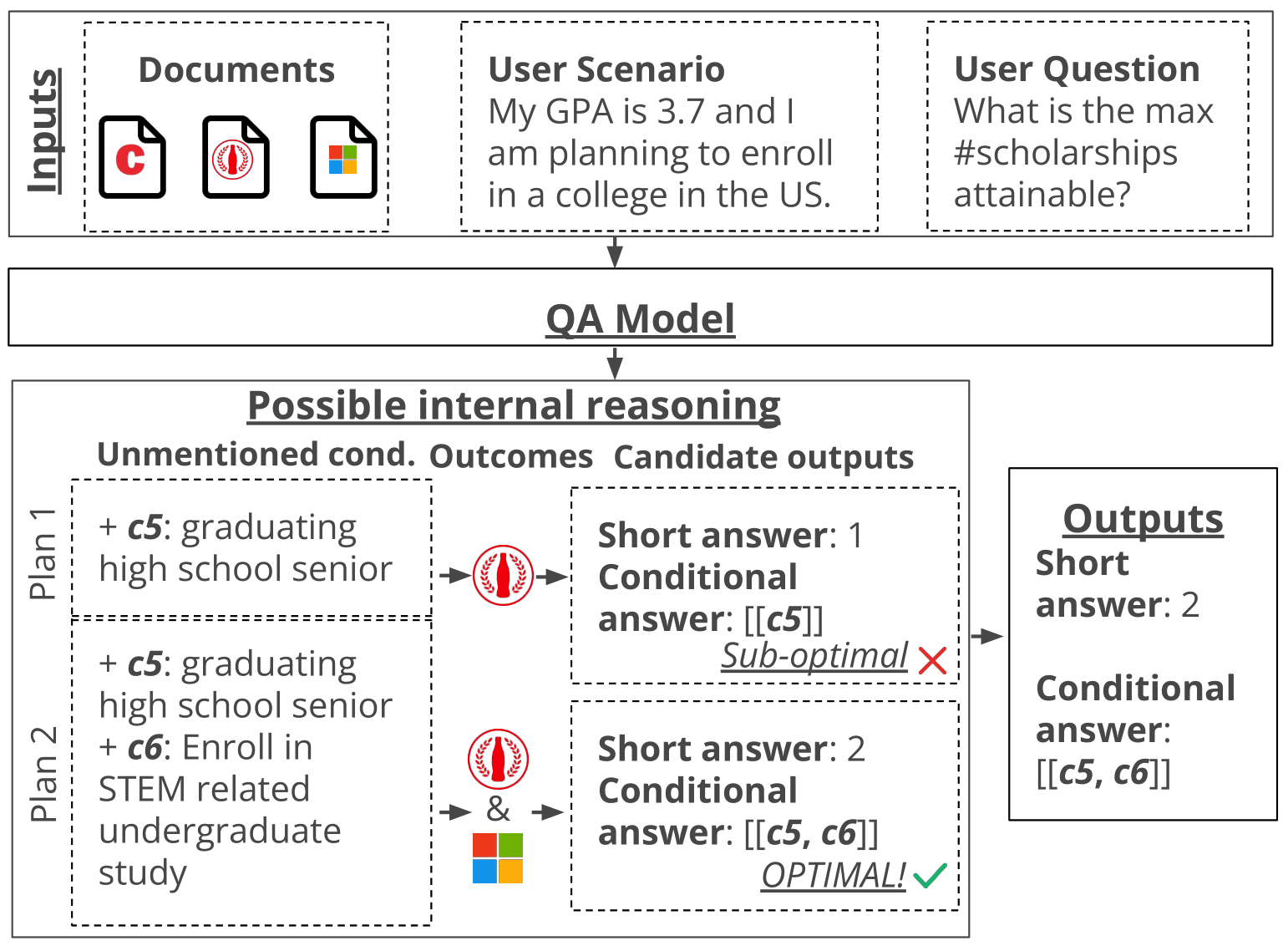}
\caption{An example of inputs, outputs and reasoning with unmentioned conditions for optimization.}
\label{fig:diagram2}
\end{figure}

Secondly, answering questions over multiple documents may require more complex reasoning capabilities for \textit{optimization}. 
Specifically, models may need to reason about a plan with additional, unmentioned conditions to achieve the best feasible outcome, based on their understanding of condition relationships and satisfiability.
Consider the user in Figure~\ref{fig:diagram2} whose objective is to maximize the number of scholarships attainable. The model could decide to use plan 1, which only leads to a sub-optimal solution, or it needs to look holistically at all three benefits, go through all possible combinations of the three scholarships (one trio, three pairs, and three singles), and leverage condition relationships to decide the group of conditions (plan 2) that results in the \textit{optimal} outcome (here, \coca{} and \ms{}). For instance, the model needs to understand the conflicting relationships between \croucher{} and both \coca{} and \ms{} to conclude that no condition group can lead to an outcome of all benefits.
Reasoning with unmentioned conditions to identify \emph{optimized} groups of conditions is not required for conditional reasoning over single documents.

Lastly, the multi-document setting magnifies the challenges in the single-document setting. In particular, it introduces more information to the context that could be relevant or irrelevant, which could potentially mislead models or lead to under-utilization, as shown in recent studies \cite{liu2024lost}, thereby making scenario and document understanding and reasoning more difficult.



To reflect the real-life challenges above, we propose a new dataset called
\dataset{} (\textbf{M}ulti-\textbf{D}ocument \textbf{C}onditional \textbf{R}easoning).
We collected documents from two domains (scholarships and jobs) and constructed questions that assess models' abilities to reason about different numbers of documents.
We benchmarked on \dataset{}, show its challenges to recent LLMs, and present insights for improvement. Most LLMs (including GPT-4o and Llama3-70B) consistently achieve around 69\% short answer accuracy and only around 40\% conditional answer F1 (around 50\% F1 for the relaxed version), demonstrating the difficulty of this task.

\section{Task Description}
In this section, we elaborate on the task of multi-document conditional reasoning, describe its inputs and outputs as well as how we evaluate the outputs.


\subsection{Inputs}
\label{sec:inputs}
The task's input consists of a set of documents, a user scenario, and a user question. 

\paragraph{Documents.} Documents in \dataset{} describe the conditions necessary to achieve an outcome,  for example,  a scholarship or a job. Typically, the \textit{outcome} of an application is the eligibility for the scholarship or the job. The  \textit{eligibility conditions} can be a set of statements whose factuality can be evaluated based on a user's scenario, for example, \textit{``be a graduating high school senior''}.
Each document can have \textsc{and}/\textsc{or} relationships among conditions.
For example, the two conditions in \croucher{} belong to \textsc{or} relationship, and the two conditions in \coca{} belong to \textsc{and} relationship.
As described in Section~\ref{sec:intro}, conditions from different documents can be related. Among these documents, conflicting, equivalent, and inclusive relationships are the most common and are sufficient to solve our tasks. Therefore, only these relationships are considered.

\paragraph{User Scenario.} 
A user scenario describes a user's background information and is self-consistent (i.e., does not include contradictory descriptions or other logical inconsistency). It consists of statements about the user's attributes regarding the application, which can include information directly relevant to the eligibility conditions or irrelevant.

\paragraph{User Questions.}
We consider three types of questions for \dataset{} to simulate what typical users would ask across multiple documents. These questions also emphasize different problem-solving skills.\\
Q1: Can I receive at least one of the outcome(s)?\\
Q2: Can I receive all the outcome(s)? \\
Q3: What is the maximum number of outcome(s) I can receive?

Among these questions, Q1 assesses models' independent reasoning capability over multiple documents (similar to single-document questions in ConditionalQA). Q2 and Q3 evaluate models' joint reasoning capability over multiple documents, which are not covered in ConditionalQA. Q3 is more challenging than Q2, because Q2 only needs to explore the unmentioned conditions to check if it can meet a fixed outcome, while Q3 further needs to find the optimal outcome by exploring different groups of unmentioned conditions.


\subsection{Outputs}
\label{sec:outputs}

Following ConditionalQA, the output of \dataset{} consists of two parts: a short answer and a conditional answer, which is a list of unmentioned conditions that need to be true for the short answer to hold.

\paragraph{Short answer.}
The short answer for Q1 and Q2 is yes/no. The short answer for Q3 is an integer value between 0 and the total number of input documents.

\paragraph{Conditional answer.}
The unmentioned conditions are those not mentioned in the user scenario but need to be satisfied for the short answer to be valid, and
there might be multiple groups of conditions that can support the short answer. In this case, although users only need to satisfy one condition group to obtain the outcome described in short answers, it is unclear which condition group(s) users can additionally satisfy. Therefore, outputs should be comprehensive to offer users as many options as possible, which means that in our task, models should output all possible condition groups.
Consider \croucher{} in our running example (Figure \ref{fig:diagram1}). It includes two conditions, i.e., $c_1$, $c_2$, with \textsc{or} relationship,  and a user can potentially be eligible for it. If these conditions are not mentioned in the user scenario, the conditional answer should include them separately and output [[$c_1$], [$c_2$]].

\subsection{Evaluation metrics}
\label{sec:metric}
We denote the outputs consisting of two parts as follows: $a$ to represent the short answer and $C$ to represent the corresponding conditional answer. Then, the output for \dataset{} is $(a, C)$. In particular, $C = \{C_1, C_2, ..., C_n\} = \{(c_1^1, ..., c_1^{l_1}), ..., (c_n^1, ..., c_n^{l_n})\}$ where $C_i$ is one possible condition group and each constituent condition $c_i^j$ (the $j$-th condition in the $i$-th condition group) need to be True for at least one group for $a$ to hold. The outputs should include all possible condition groups.
We developed metrics for each part separately: the accuracy for the short answer
and the retrieval performance for the conditional answer.

\paragraph{Accuracy of short answers.} The accuracy
is defined as the exact match score of the predicted short answer and the gold short answer.

\paragraph{Performance of conditional answers.}
For each short answer, we labeled the required condition groups as the gold conditional answer. Therefore, we can compute precision, recall, and F1 of gold conditions to evaluate the conditional answers. We considered two different evaluation setups. 
Under a strict setup, an order-insensitive exact match of all required condition groups is considered correct. In this case, each condition group is treated as one unit for comparison. Consider the optimal conditional answer shown in Figure \ref{fig:diagram2} [[$c_5$, $c_6$]]. Under the strict setup, a prediction of [[$c_6$, $c_5$]] has a F1 of 1 but [[$c_5$, $c_6$, $c_7$]] has a F1 of 0.
We also considered a relaxed setup, using a metric that allows partial matches to the gold conditional answer (e.g., [$c_5$, $c_6$, $c_7$] partially matches [$c_5$, $c_6$] and relaxed F1 is 0.8).
The core idea is to compute a 1-1 mapping between the condition groups of the prediction and gold outputs that maximize the sum of F1 of condition groups in the mapping, which is detected by an off-the-shelf solver \footnote{We used scipy.optimize.linear\_sum\_assignment.}. We then computed the average precision, recall, and F1 for each prediction and gold condition group to obtain the overall precision, recall, and F1.






\section{Data collection}
In this section, we provide an overview of the data collection process. More details are provided in Appendix \ref{app:datasets}.

\subsection{Corpus}

We collected HTML documents from two domains: scholarships and jobs. Documents from these domains typically include eligibility conditions, and conditions from different documents are likely to have overlapping attributes (e.g., GPA, degree, majors, years of experience). As discussed in Section \ref{sec:inputs}, relationships of these conditions can be categorized into conflicting, equivalent, or inclusive.
Similar to ConditionalQA, we retained most HTML tags and page content except common sections and irrelevant information (e.g., navigation bar).

\subsection{Human annotations}
\label{sec:annotations}
As mentioned in Section \ref{sec:inputs}, inputs consist of three components: documents, user scenarios, and user questions.
In this section, we discuss the annotations collected for documents. Appendix \ref{app:annotation-interfaces} shows the web interfaces annotators used. For each document, there are three annotation tasks:\\
Task 1: extracting conditions from documents. Outputs are the HTML sentences that describe eligibility conditions (e.g., $c_1, c_2$ in Figure \ref{fig:diagram1}).\\
Task 2: identifying \textsc{and}/\textsc{or} relationships of conditions \textit{within} documents. Outputs are boolean expressions over documents' conditions (e.g., $c_1$ \textsc{or} $c_2$).\\
Task 3: labeling condition relationships \textit{across} documents as conflicting, equivalent, or inclusive (e.g., $c_1$ conflicts with $c_3$).

\subsection{Scenario Generation}
\label{sec:scenario}
Using annotations discussed above, user scenarios were constructed based on a randomly sampled group of documents (2-5 in our datasets) to mimic users' background information. Scenarios were designed to include both information directly relevant to the conditions mentioned in the document group (and thus implies the satisfiability of these conditions) and irrelevant information. 


We started with sampling conditions. One to five conditions were sampled from each document in the group as relevant information. From the other documents in the corpus, one to ten conditions were sampled as irrelevant information. Then, we assigned True/False values to each sampled condition to indicate their satisfiability in the user scenario. We leveraged condition relationships during the sampling and value assignment process (e.g., avoid sampling conflicting conditions and avoid assigning reverse values to equivalent conditions) to ensure logical consistencies.  Lastly, we used an LLM (GPT-4 in our case)
to generate scenarios based on the sampled conditions and their value assignment. Annotators further verified the consistency of these scenarios.
To ensure a diversity of scenarios, we generated 5 scenarios for each group of documents.
We further avoided generating \textit{too many} simple scenarios whose solving processes do not require reasoning with condition relationships and unmentioned conditions, as these are the key characteristics of our datasets.
Details are provided in Appendix \ref{app:easy-scenarios}.

\subsection{Gold answer generation}
\label{sec:mapping}
We discuss how to use annotations collected in Section \ref{sec:annotations} to determine gold answers automatically. The high-level idea is to represent the problem as a boolean satisfiability problem, which can then be solved with off-the-shelf symbolic solvers \footnote{We used https://github.com/cjdrake/pyeda.}.

Given the documents and user scenarios in the inputs, a joint boolean expression was constructed by merging each document's boolean expression.
The values of conditions in the joint expression were assigned based on their corresponding values (True/False) in user scenarios, while considering the value constraints implied by condition relationships.
The joint boolean expression was then solved symbolically to obtain (1) the feasibility of obtaining all benefits (2) all groups of unmentioned conditions to support the feasibility. We leveraged condition relationships to post-process these outputs to ensure logical consistency and perform simplification. Solutions to the three question types were obtained by running this process.









\subsection{Statistics}
We collected 20 documents from each domain, thus 40 documents in total. From all possible groups of 2-5 documents, we sampled 55, 68, 31, and 10 groups of 2, 3, 4, and 5 documents, respectively, to construct scenarios.
The scenario generation process resulted in 518 and 620 scenarios for the scholarships and internships domains, respectively, a total of 1138 scenarios. There are three question types per scenario, and thus 3414 questions.
Of these, 961 questions (28.1\%) require an understanding of condition relationships to arrive at the correct answer. In other words, answers are different if condition relationships are not considered. 2250 questions (66.0\%) include non-empty conditional answers in their outputs.
Each task was performed by several annotators. Due to little ambiguity, all labels obtained a majority agreement (with Fleiss' kappa \cite{fleiss1971measuring} above 0.7).

\section{Benchmark}
\label{sec:benchmark}

In this section, we evaluate recent LLMs on our dataset and elaborate on their performance under different settings.

\subsection{Experimental setup}
\label{sec:setup}

We used LLMs of various sizes, including GPT-4o \cite{achiam2023gpt}, Llama-3-Instruct (70B and 8B) \cite{touvron2023llama}, and Gemma-7B-Instruct \cite{team2024gemma}. For prompting strategies, we adopted 0-shot prompting, 1-shot\footnote{We did not provide more examples due to input size limit.}, and CoT  with 1-shot example \cite{wei2022chain} (we did not provide human-crafted reasoning steps for the 1-shot example because there are multiple valid ways to reason and we avoid restricting models to any particular approach). The temperature was set to 0 to minimize randomness. We did not do any post-processing for positive short answers. However, if the short answer is ``no'' or 0, subsequent generated conditional answers were ignored to avoid potential inconsistency in the generated outputs \footnote{We observed cases where LLMs did not follow the instruction and generated conditional answers for a short answer ``no'' or 0.}. 

\paragraph{Methods.} The baseline method refers to the standard end-to-end setup where prompts that include task instruction, inputs, and information from prompting strategies are given to LLMs to generate outputs. In section \ref{sec:understanding}, we added various hints to the baseline prompts. Prompts of all methods are provided in Appendix \ref{app:prompts}.

\paragraph{Single-document datasets.} To demonstrate the challenges introduced by the multi-document setting, 
we constructed a single-document setting based on our dataset for comparison. 
Specifically, we used the same collection of documents and user questions, including Q1, Q2, and Q3. 
Rather than including all the documents associated with a pair of user scenarios and questions in inputs, we used only one document in the single-document setting. 
We then updated scenarios by only including the relevant conditions from the selected single document and the same set of irrelevant conditions sampled for the multi-document setting.
Since we can construct multiple single-document data instances for each instance in our multi-document dataset, we did a down-sampling for a fair comparison.


\begin{table*}
\centering

\begin{adjustbox}{max width=\linewidth}
\begin{tabular}{cccc|ccc|ccc|ccc|ccc}
& \multicolumn{3}{c}{Baseline} & \multicolumn{3}{c}{\circlednum{a} With document conditions} & \multicolumn{3}{c}{\circlednum{b} With condition satisfiability} & \multicolumn{3}{c}{\circlednum{c} With condition relationships} & \multicolumn{3}{c}{With \circlednum{a} + \circlednum{b} + \circlednum{c}}\\
\cmidrule(lr){2-4} \cmidrule(lr){5-7} \cmidrule(lr){8-10} \cmidrule(lr){11-13} \cmidrule(lr){14-16}
& \multicolumn{1}{c}{Answer} & \multicolumn{2}{c}{Conditional answer F1} & \multicolumn{1}{c}{Answer} & \multicolumn{2}{c}{Conditional answer F1} & \multicolumn{1}{c}{Answer} & \multicolumn{2}{c}{Conditional answer F1} & \multicolumn{1}{c}{Answer} & \multicolumn{2}{c}{Conditional answer F1} & \multicolumn{1}{c}{Answer} & \multicolumn{2}{c}{Conditional answer F1} \\
\cmidrule(lr){2-2} \cmidrule(lr){3-4} \cmidrule(lr){5-5} \cmidrule(lr){6-7} \cmidrule(lr){8-8} \cmidrule(lr){9-10} \cmidrule(lr){11-11} \cmidrule(lr){12-13} \cmidrule(lr){14-14} \cmidrule(lr){15-16}
& Accuracy & Exact & Relaxed & Accuracy & Exact & Relaxed & Accuracy & Exact & Relaxed & Accuracy & Exact & Relaxed & Accuracy & Exact & Relaxed \\
\midrule
0-shot \\
\midrule

GPT-4o & 76.4 & 46.8 & 60.8 & 76.4 & 56.1 & 66.2 & 79.5 & 52.5 & 66.3 & 81.2 & 48.5 & 62.8 & 86.3 & 57.2 & 69.4\\
Llama-3-70B-Instruct & 59.2 & 19.5 & 40.0 & 66.6 & 27.7 & 47.2 & 60.7 & 25.9 & 43.4 & 57.0 & 16.6 & 33.5 & 73.0 & 38.8 & 55.7\\
Llama-3-8B-Instruct & 23.4 & 3.01 & 11.3 & 51.4 & 13.4 & 21.0 & 23.8 & 4.82 & 11.8 & 23.7 & 3.29 & 13.8 & 61.6 & 35.7 & 43.4\\
Gemma-7B-Instruct & 52.5 & 14.6 & 25.4 & 62.2 & 17.5 & 27.7 & 58.1 & 14.1 & 18.5 & 54.2 & 14.2 & 27.2 & 46.4 & 8.39 & 19.3\\

\midrule
1-shot \\
\midrule
GPT-4o & 68.7 & 40.4 & 54.2 & 68.3 & 46.2 & 58.3 & 72.6 & 47.4 & 60.9 & 75.2 & 39.1 & 53.0 & 82.1 & 53.0 & 66.7\\
Llama-3-70B-Instruct & 68.2 & 33.0 & 48.9 & 71.0 & 36.8 & 54.9 & 70.4 & 38.9 & 53.3 & 72.4 & 31.8 & 45.6 & 72.9 & 41.9 & 57.9\\
Llama-3-8B-Instruct & 58.7 & 27.1 & 41.6 & 64.4 & 27.1 & 43.5 & 60.7 & 29.6 & 42.6 & 63.8 & 28.4 & 42.6 & 63.2 & 31.6 & 45.7\\
Gemma-7B-Instruct & 58.9 & 23.6 & 34.6 & 59.9 & 27.5 & 39.6 & 62.8 & 21.9 & 27.8 & 67.6 & 25.7 & 38.3 & 74.5 & 25.3 & 38.6\\

\midrule

1-shot w/ COT  \\

\midrule

GPT-4o & 67.8 & 46.8 & 59.4 & 70.9 & 52.4 & 64.7 & 76.4 & 53.0 & 67.3 & 75.8 & 47.3 & 59.9 & 87.9 & 63.2 & 78.0\\
Llama-3-70B-Instruct & 63.6 & 30.9 & 45.7 & 65.6 & 37.5 & 52.5 & 63.3 & 32.7 & 45.6 & 64.8 & 27.4 & 41.0 & 71.7 & 40.7 & 57.1\\
Llama-3-8B-Instruct & 61.3 & 29.3 & 41.0 & 64.6 & 28.6 & 39.4 & 63.8 & 30.2 & 37.4 & 63.4 & 30.2 & 40.5 & 67.0 & 31.0 & 40.7\\
Gemma-7B-Instruct & 59.0 & 18.6 & 29.4 & 55.9 & 18.9 & 33.2 & 61.8 & 17.8 & 25.7 & 63.9 & 20.1 & 32.4 & 68.7 & 26.0 & 37.6\\

\bottomrule
\end{tabular}
\end{adjustbox}

\caption{Baseline performance on short and conditional answers
is poor, but performance increases as more hints are provided in prompts to models (e.g., GPT-4o has short answer accuracy of 86.3\% with \circlednum{a}+\circlednum{b}+\circlednum{c}, higher than 79.5\% with \circlednum{b}, and 76.4\% with baseline). However, performance is still imperfect when all hints are given.}
\label{tab:gold}
\end{table*}

\subsection{Baseline performance}
We first provide an overview of baseline performance on our dataset, using metrics defined in Section \ref{sec:metric}. Then, we investigate the performance under different question types varying the number of input documents to introduce the challenges for LLMs, such as long-context understanding and conditional reasoning with optimization.

\paragraph{In general, current LLMs struggle with conditional reasoning over multiple documents.}
For baseline performance (column 1 in Table \ref{tab:gold}), although the short answer accuracy of the best-performing model (GPT-4o) is around 75 \%, it did not exceed 69\% for the majority of the models. Furthermore, the relaxed conditional answer F1 is only around 50\%, and the exact F1 is even lower (around 40\%). These numbers show that the overall baseline performance on this task is low across all models and prompting strategies, suggesting the challenging nature of \dataset{}.
The short answer accuracy is significantly higher than the conditional answer F1, suggesting that reasoning about conditional answers is much more difficult than reasoning about short answers. The low relaxed F1 also demonstrates that while models can generate some partially correct groups of conditions, they still perform poorly to identify the complete groups. This is potentially because LLMs may be better at identifying signals to make quick decisions (for short answers), while struggling with reasoning step by step and considering all possibilities of unmentioned conditions for conditional answers.

\begin{table}
\centering

\begin{adjustbox}{max width=\linewidth}

\begin{tabular}{ccc|cc|cc|cc}
& \multicolumn{2}{c}{Q1 (at least one)} & \multicolumn{2}{c}{Q2 (all)} & \multicolumn{2}{c}{Q3 (max number)} & \multicolumn{2}{c}{Overall (Q1, Q2, Q3)} \\
\cmidrule(lr){2-3} \cmidrule(lr){4-5} \cmidrule(lr){6-7} \cmidrule(lr){8-9} 
& \multicolumn{1}{c}{Answer} & \multicolumn{1}{c}{Cond.} & \multicolumn{1}{c}{Answer} & \multicolumn{1}{c}{Cond.} & \multicolumn{1}{c}{Answer} & \multicolumn{1}{c}{Cond.} & \multicolumn{1}{c}{Answer} & \multicolumn{1}{c}{Cond.}\\
\cmidrule(lr){2-2} \cmidrule(lr){3-3} \cmidrule(lr){4-4} \cmidrule(lr){5-5}  \cmidrule(lr){6-6}  \cmidrule(lr){7-7}  \cmidrule(lr){8-8}  \cmidrule(lr){9-9}
& Accuracy  & Exact F1 &  Accuracy & Exact F1 & Accuracy & Exact F1& Accuracy & Exact F1\\
\midrule
Multiple documents \\
\midrule
GPT-4o & 66.1 & 24.0 & 82.1 & 76.9 & 56.2 & 17.9 & 68.7 & 40.4\\
Llama-3-70B-It & 75.9 & 14.0 & 81.0 & 76.4 & 43.3 & 5.82 & 68.2 & 33.0\\
Llama-3-8B-It & 62.6 & 9.0 & 80.3 & 79.0 & 37.6 & 4.04 & 58.7 & 27.1\\
Gemma-7B-It & 59.3 & 12.5 & 78.2 & 76.5 & 47.2 & 3.6 & 58.9 & 23.6\\

\midrule
Single document \\
\midrule
GPT-4o & 81.0 & 57.8 & 81.2 & 60.1 & 88.5 & 68.5 & 83.5 & 62.0\\
Llama-3-70B-It & 83.4 & 40.9 & 80.5 & 39.6 & 85.4 & 40.5 & 83.1 & 40.4\\
Llama-3-8B-It & 50.4 & 25.8 & 50.8 & 26.2 & 63.1 & 24.2 & 54.8 & 25.4\\
Gemma-7B-It & 72.8 & 27.5 & 53.6 & 25.8 & 78.8 & 25.1 & 68.4 & 26.1\\

\bottomrule
\end{tabular}

\end{adjustbox}
\caption{1-shot baseline performance under different question types and number of documents. Questions (e.g., Q3) requiring more complex reasoning are generally more challenging. Longer contexts lead to additional complexity that contributes to lower performance.}
\label{tab:q_type}
\end{table}

\paragraph{LLMs struggle more with conditional reasoning requiring optimization.}
We further examine the performance of different question types to understand the impacts of reasoning complexity on final performance.
As shown in the top half of Table \ref{tab:q_type}, under the multi-document setting, across all models, the average short answer accuracy is 46.1\% for Q3, 66.0\% for Q1, and 80.4\% for Q2; the average conditional answer F1 is 7.84\% for Q3, 14.9\% for Q1, and 77.2\% for Q2.
These numbers show that questions (e.g., Q3) that require exploring condition groups from multiple documents \textit{jointly} to detect the \textit{optimal} one are more difficult compared to questions (e.g., Q1) that only require reasoning over documents \textit{independently}.
However, we observe Q2 performs better than Q1, although it involves reasoning with conditions over multiple documents.
This is potentially because the model might be able to answer Q2 through some shortcut signals, so the question is somehow simplified. For example, an unsatisfied critical condition or a pair of conflicting conditions make it impossible to obtain all benefits.
In this case, the model may not do reasoning step by step and consider all possibilities of unmentioned conditions as we expected.


\paragraph{In general, more documents increase context length, which increases the difficulty.}
As seen in Table \ref{tab:q_type}, across all models,
performance drops significantly for both Q1 and Q3, on average 5.9\% and 32.9\% for short answer accuracy and 23.1\% and 31.7\% for conditional answer F1, respectively, going from the single-document setting to the multi-document setting.
As mentioned in Section \ref{sec:setup}, comparing the single-document and the multi-document datasets, the same questions and similar user scenarios were used; the number of documents is the primary difference. Thus, the lowered performance suggests that \textit{longer contexts} due to more documents indeed increase the difficulty.
However, for Q2, both short answer accuracy and conditional answer F1 increase. As explained in the last paragraph, identifying any unsatisfied critical conditions or conflicting conditions may simplify the process of answering Q2. The multi-document setting may increase the chance that these shortcuts appear.

\subsection{Analysis on Condition Understanding} 
\label{sec:understanding}


As mentioned in Section \ref{sec:intro}, correctly understanding conditions and condition relationships are critical to solve \dataset{}. Therefore, in this section, we analyze the performance by examining the impact of models' understanding of conditions. In terms of condition understanding, models need to be able to \circlednum{a} perform condition extraction (i.e., identifying the eligibility conditions mentioned in the document and their \textsc{and}/\textsc{or} relationships) \circlednum{b} identify the satisfiability of conditions according to user scenarios \circlednum{c} identify the relationships of conditions across documents. 
We added the gold information of \circlednum{a}, \circlednum{b}, and \circlednum{c} to the baseline method to examine their impact on LLMs' performance. Results are reported in columns 2-5 in  Table~\ref{tab:gold}.
\paragraph{Providing gold information of \textit{each} understanding component significantly increases LLM's performance.} By comparing columns 2-4 in Table \ref{tab:gold} with the baseline performance in column 1, we observe that across all models and prompt strategies, models perform 4.96\% and 4.67\% better with \circlednum{a} and 3.02\% and 2.93\% better with \circlednum{b}
on short answer accuracy and conditional answer F1, respectively. These results demonstrate that models can effectively leverage these hints to help with reasoning but potentially illustrate the limited understanding capabilities of LLMs due to the poor baseline performance.
While providing \circlednum{c} condition relationships leads to an average of 3.77\% higher accuracy for short answers, the average conditional answer F1 remains the same. From error analysis, we observe that models understand the given hints but have limited reasoning ability to use them correctly. More details are provided in Appendix \ref{app:low-f1}.

\paragraph{Providing gold information of \textit{all} understanding components achieves the highest, but not perfect, performance.}
Comparing the last column of Table \ref{tab:gold} and columns 1-4 shows that, across all models and prompt strategies, models achieve the best performance when hints from all understanding components are provided, having an average increase of 11.5\% and 9.93\% for short and conditional answers compared to the baseline. These results demonstrate that models can effectively combine hints of multiple types. However, the imperfect scores suggest limited reasoning capabilities.



\begin{table*}
\centering
\begin{adjustbox}{max width=\linewidth}
\begin{tabular}{cccccc}
Error types & \% (Baseline) &  \% (gold) & Setting & Correct answers & Predictions\\
\midrule
\multicolumn{6}{l}{Short answer} \\
\midrule
\makecell{Over-reaction to negative signals}  & 22 & 0 & \makecell{Number of documents: 2 \\ Condition satisfiability: False conditions for \\ some benefits only (and otherwise True) \\ Have conflicting conditions: Yes} & Eligible for at least one benefit & Eligible for no benefit\\
\cdashline{1-6}
\makecell{Incorrect association to negative signals} & 31 & 7 & \makecell{Number of documents: 2 \\ Condition satisfiability: all True \\ Have conflicting conditions: No} & Eligible for at most 2 benefits & Eligible for at most 1 benefit \\
\cdashline{1-6}
\makecell{Incorrectly handling conflicting \\ relationships or signals} & 47 & 22 & \makecell{Number of documents: 2 \\ Condition satisfiability: some False \\ Have conflicting conditions: Yes} & Eligible for at most 1 benefit & Eligible for at most 2 benefits\\
\midrule
\multicolumn{6}{l}{Conditional answer} \\
\midrule
Incompleteness & 62 & 38 & doc19-c13 is unmentioned in scenario & [[doc19-c11, doc19-c12, doc19-c10, \textbf{doc19-c13}]] & [[doc19-c11, doc19-c12, doc19-c10]]\\
Redundancies & 27 & 9 & doc12-c3 is satisfied & [[doc12-c1]] & [[doc12-c1, doc12-c3]] \\
\makecell{Include conditions from ineligible benefits} & 30 & 31 & User is only eligible for documents 5 and 12 & [[doc5-c34], [doc12-c1]] & [[doc5-c34], [doc12-c1], \textbf{[doc19-c9]}] \\
Merge condition groups & 6 & 7 & doc19-c10 conflicts with doc5-c34 & [[doc5-c34, doc5-c35], [doc19-c10, doc19-c11]] &[[doc5-c34, doc5-c35, doc19-c10, doc19-c11]] \\
Split condition groups & 12 & 15 & User is eligible for both document 2 and 7 & [[doc2-c16, doc7-c1, doc7-c2]] & [[doc2-c16], [doc7-c1, doc7-c2]]\\
Include irrelevant conditions & 5 & 0 & doc13-c16 is not an eligibility condition & [[doc13-c9]] & [[doc13-c9, doc13-c16]] \\
\bottomrule
\end{tabular}
\end{adjustbox}

\caption{Common error types for GPT-4o 0-shot and Llama-70B 1-shot, showing limited understanding and reasoning abilities. Providing gold information significantly reduces the overall error rate.}
\label{tab:error_analysis}
\end{table*}





\section{Error analysis}
\label{sec:error}





Section \ref{sec:benchmark} describes the challenges in \dataset{} introduced in Section \ref{sec:intro}, including longer context and reasoning complexity of different types of questions, and hints at models' limited capabilities of performing this task.
In this section, we perform a detailed error analysis to better understand the error sources.
To achieve it, we randomly sampled 100 questions from the two best-performing models (GPT-4o 0-shot and Llama3-70B-Instruct 1-shot). Categorized error types are listed in Table \ref{tab:error_analysis}. Overall, we find that the total error rate is much lower when gold information about conditions is provided. This aligns with our observation in Table \ref{tab:gold} that models can leverage hints to help this task. In the rest of this section, we will focus on the errors happening when running the baseline without gold information provided.



\subsection{Short answers analysis}
As seen in the top half of Table \ref{tab:error_analysis}, models made three major mistakes on short answers. Firstly, \textit{they tend to overreact to negative signals}. We observe that in many cases, although there are descriptions of unsatisfied conditions for only \textit{some} documents in the user scenario, models conclude that users are \textit{not} eligible for \textit{any} benefits immediately. In user scenarios, descriptions of negative signals usually start with words such as \textit{``However''} or \textit{``I am not''}.
We observe that models can be misled by such specific wordings and fail to logically reason about the actual satisfiability of conditions (e.g., conditions that \textit{are satisfied}) to generate correct outputs.

Secondly, \textit{models can make a wrong association to negative signals} and conclude that users are eligible for fewer benefits than actually attainable. Specifically, models could misinterpret the satisfiability of a user fact. For instance, given the user scenario \textit{``my research is in the field of Computer Science''}, GPT-4o concluded that the user did not satisfy the condition \textit{``declare a Biochemistry and/or Molecular Biology major, or related discipline''} and thus the user is ineligible for this scholarship. Yet, the user fact and the condition are not contradictory, and probably indicate a wrong association produced by the model. 

Finally, \textit{models can fail to leverage conflicting relationships or make a wrong association to conflicting signals} and conclude that users are eligible for more benefits. 
For instance, given the user scenario \textit{``I have a Bachelor's degree in Civil Engineering''}, GPT-4o concluded that the user could potentially satisfy the condition \textit{``You are an undergraduate student ... with a major in Environmental Studies ...''} (denoted as $c_a$) and included it in the conditional answer. However, the conclusion made by GPT-4o actually contradicts the user fact (once one has a Bachelor's degree, they are no longer an undergraduate student.). 
Models can also fail to identify conflicts. For the same example above, GPT-4o was unable to identify that $c_a$ conflicts with the condition \textit{``Bachelor's degree (or higher) in Civil, Mechanical, or Architectural Engineering ...''} either. Due to both misinterpretations, models concluded that the user was eligible for an internship when they were ineligible.

These behaviors suggest that models struggle to understand condition satisfiability and relationships and incorporate them to reason logically.

\subsection{Conditional answers analysis}

As seen in the bottom half of Table \ref{tab:error_analysis}, models made three major mistakes on producing conditional answers. First, \textit{conditional answers can be incomplete}, meaning that models miss some unmentioned conditions that are critical to support the short answer in the output. 
Since the error rate decreases after gold information of \circlednum{a} and \circlednum{b} is provided, we think this error happens probably because models, initially, failed to recognize eligibility conditions from documents or conditions whose satisfiability is not implied by user scenarios. It implies that understanding document conditions and condition satisfiability based on user scenarios remains challenging for models.

Secondly, \textit{outputs can be redundant} as models repeat conditions that are already mentioned or satisfied according to user scenarios. It suggests that models may have difficulty in identifying exactly which conditions have been satisfied and which have not been satisfied.

Lastly, \textit{models can include conditions from ineligible documents}, which suggests their limited capability to ensure logical consistency between short and conditional answers in the outputs.

Overall, the error analysis highlights that \textbf{extracting document conditions and understanding condition satisfiability based on user scenarios considering condition relationship are big challenges} for models to solve \dataset{}. \textbf{Models also often fail to perform solid logical reasoning and ensure logical consistency in the outputs.}




\section{Conclusion}
Conditional reasoning is crucial in many domains, such as scholarships, job applications, and government benefits. It involves understanding the eligibility conditions and determining \emph{optimal} outcomes based on users' satisfiability of conditions. However, existing works focus only on such reasoning over single documents, neglecting situations where users want to search for \textit{optimized} outcomes that span through multiple documents. In this paper, we introduced \dataset{} to address this gap and benchmarked recent large language models on this dataset. As evidenced by the results, the multi-document setting brings significant challenges to models and suggests their limited understanding and reasoning capabilities. We hope this serves as the foundation for future work that examines complex conditional reasoning.

\clearpage
\section{Ethics}
We recruited six graduate volunteers to perform the annotations. They were given an onboarding process to familiarize themselves with the tasks and were also invited to a group chat to discuss any unclear examples. Because this dataset involves only factual annotations, no subjective opinions or personal information were collected, and thus, it should pose minimal risks to annotators and the general public. We ensure fair compensation for each volunteer, considering the minimum salary of the region these volunteers are in.
Our institution's ethical committee reviews this work and has determined it to be exempt. We abide by the policies required by the institution throughout the data collection process. 

\section{Limitations}
This work mainly focuses on domains of scholarship and job applications, and explores models' performance of answering typical questions in these domains that require reasoning over multiple documents. Though we believe it is a good test bed for models' conditional reasoning capability, there are more domains and possible questions available in reality. We believe extending the domains of documents and types of questions can be a promising future work to continue the exploration of this direction.
Furthermore, as described in Section~\ref{sec:intro}, understanding condition relationships across documents is critical for answering multi-document questions. However, this may require beyond commonsense knowledge. For example, eligibility conditions might include requirements on citizenship, and some countries allow dual citizenship while others do not. In this work, we did not dive deep into exploring whether the model may have such knowledge, and how much it would influence a model's reasoning capability. We think investigating models' conditional reasoning capability with external knowledge would also be an interesting future work.

\bibliography{custom}

\appendix


\section{Datasets}
\label{app:datasets}

\subsection{Human annotations}
\label{app:annotation-details}
We provide more details for the process of annotating documents.
As described in Section \ref{sec:annotations}, for each document, we have three annotation tasks: (1) extracting conditions from documents, (2) identifying \textsc{and}/\textsc{or} relationships of conditions within a document, and (3) labeling the relationship between conditions across documents.

Specifically, human annotators were asked to identify the eligibility conditions that must be satisfied (we ignore conditions that are not strictly required such as ``preferred qualifications'') as well as the boolean relationships (\textsc{and}/\textsc{or}) of these conditions. Some sentences in the HTML documents might include multiple conditions and if these conditions are separated by punctuations (e.g., period) or conjunctions (e.g., words and/or), annotators further splitted them into self-contained conditions and provided a mapping between the HTML sentence and the extracted conditions. The above annotations give us a boolean expression over the document's conditions.

After each document's conditions were extracted, the relationship for each pair of conditions was labeled as conflicting (one condition being True means the other is False), equivalent (the two conditions have the same satisfiability status), or inclusive (the subset condition being True means the superset condition is True and the superset condition is False means the subset condition is False).

\subsection{Scenario generation}
\label{app:easy-scenarios}
We provide more details on how to prevent over-generating simple scenarios. Since the goal of \dataset{} is to investigate models' abilities to reason with condition relationships and unmentioned conditions, having \textit{too many} simple questions whose solving processes do not require such capabilities defeats the purpose of our dataset.
To achieve this, we assigned two sets of values for the same set of sampled conditions: one to be all True and the other to True/False = 0.8/0.2. These strategies decrease the chance that unsatisfied critical conditions appear, so models are less likely to be able to conclude answers directly from condition satisfiability implied from user scenarios and thus will have to further leverage condition relationships during reasoning. In addition, because critical conditions are more likely to be satisfied, short answers are more likely to be positive and include non-empty conditional answers, which require capabilities to reason with unmentioned conditions.


\subsection{Gold answer generation}
We provide more details for generating gold answers.
Once outputs were obtained by running symbolic solvers of the joint boolean expression (discussed in Section \ref{sec:mapping}), we performed post-processing on these outputs to (1) remove conditional answers that involve conflicting conditions (and adjust the short answer accordingly) and (2) simplify conditional answers to keep one of the two equivalent conditions and the subset condition of an inclusive relationship.
We ran the above process once to solve Q2 (all), once for each document to solve Q1 (at least one), and once for all possible document combinations to obtain the optimal combination to solve Q3. As discussed in Section \ref{sec:annotations}, a mapping exists from the HTML sentences to the conditions. Therefore, as a final step, we map the conditions back to the HTML sentences to obtain the final gold answer.

\subsection{Annotation interfaces}
\label{app:annotation-interfaces}
\begin{figure*}
\centering
\includegraphics[width=\textwidth]{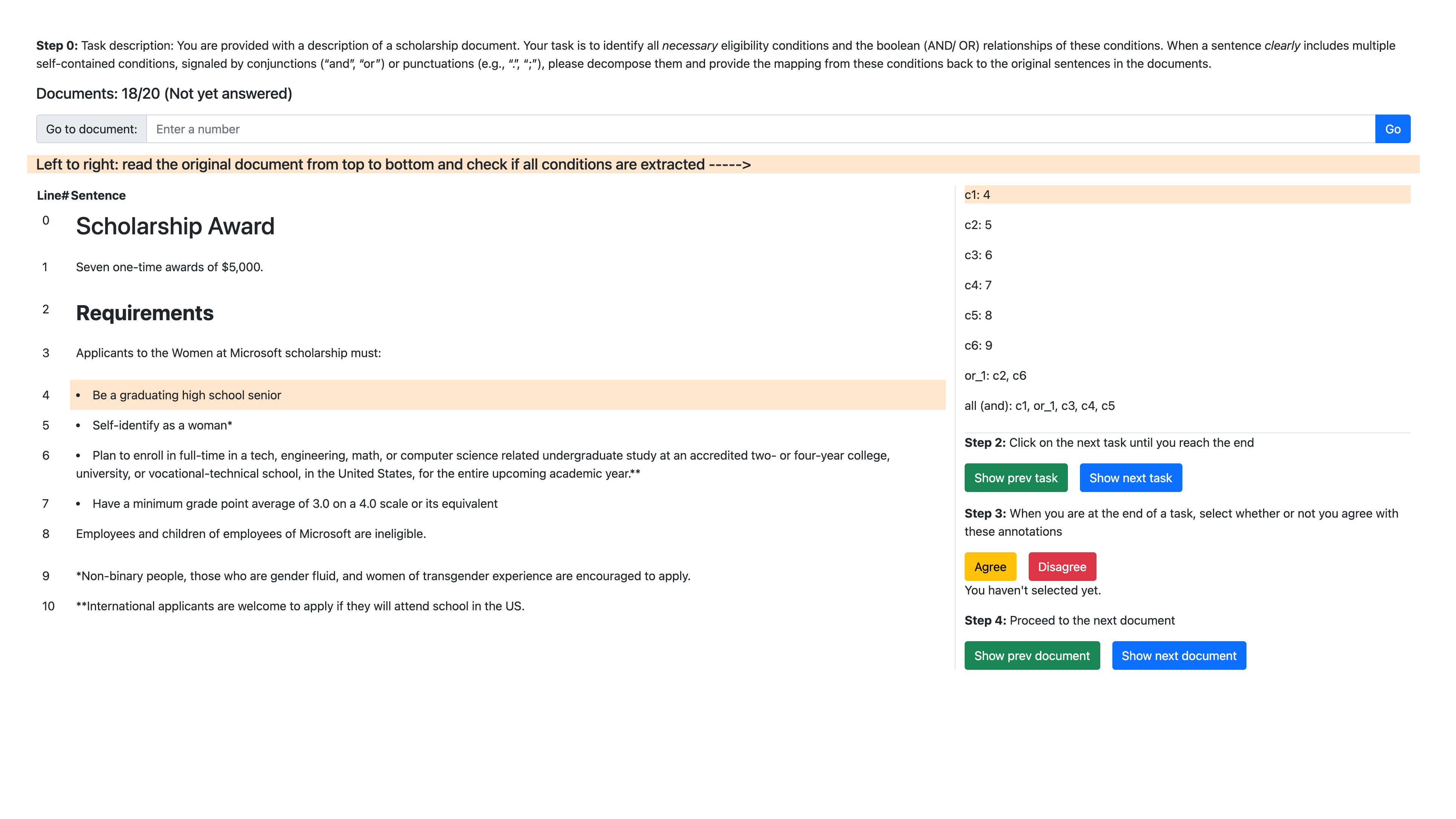}
\caption{Interface for condition extraction.}
\label{fig:annotation_doc_1}
\end{figure*}

\begin{figure*}
\centering
\includegraphics[width=\textwidth]{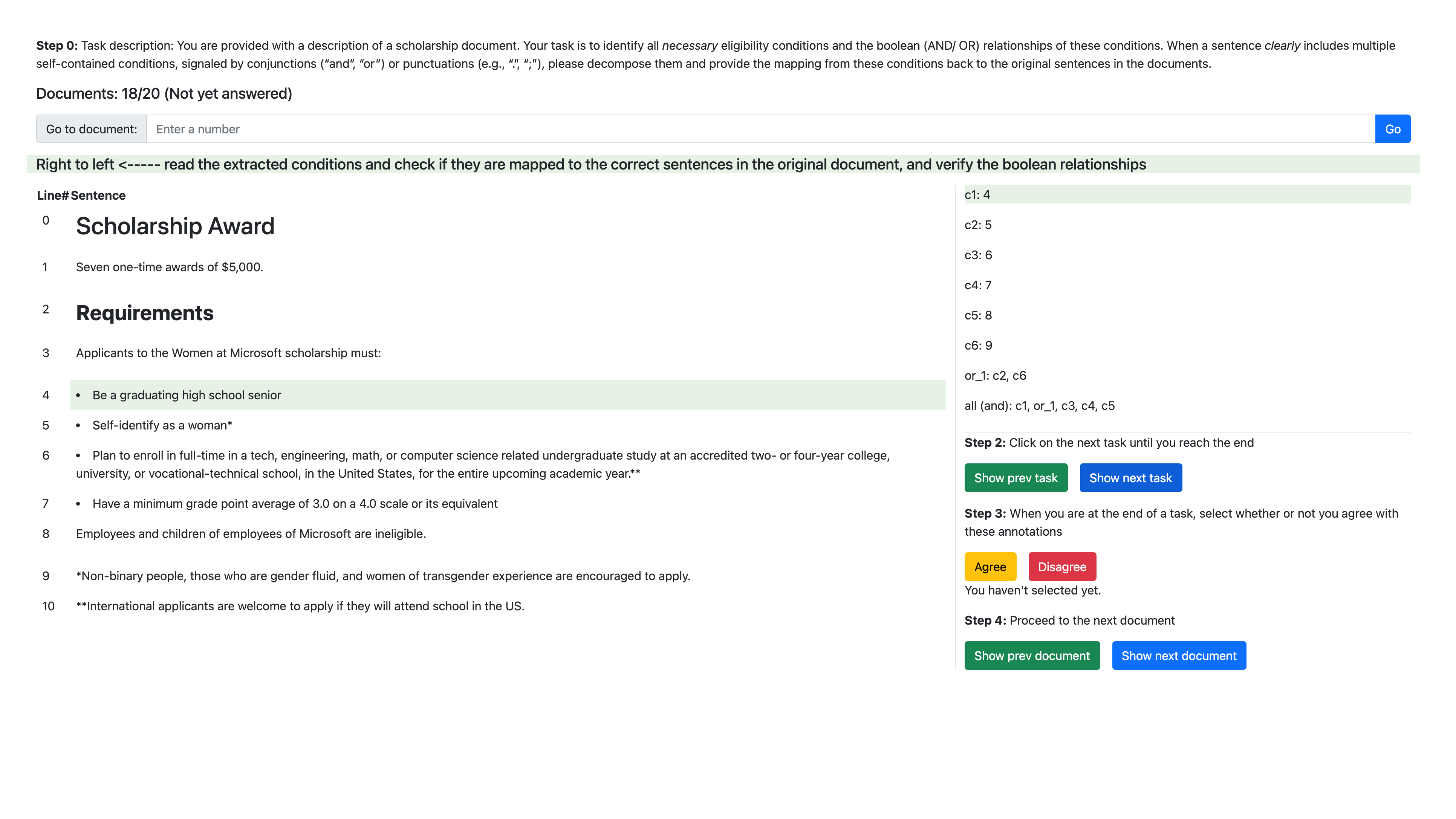}
\caption{Interface for identifying mapping and boolean relationships.}
\label{fig:annotation_doc_2}
\end{figure*}

\begin{figure*}
\centering
\includegraphics[width=\textwidth]{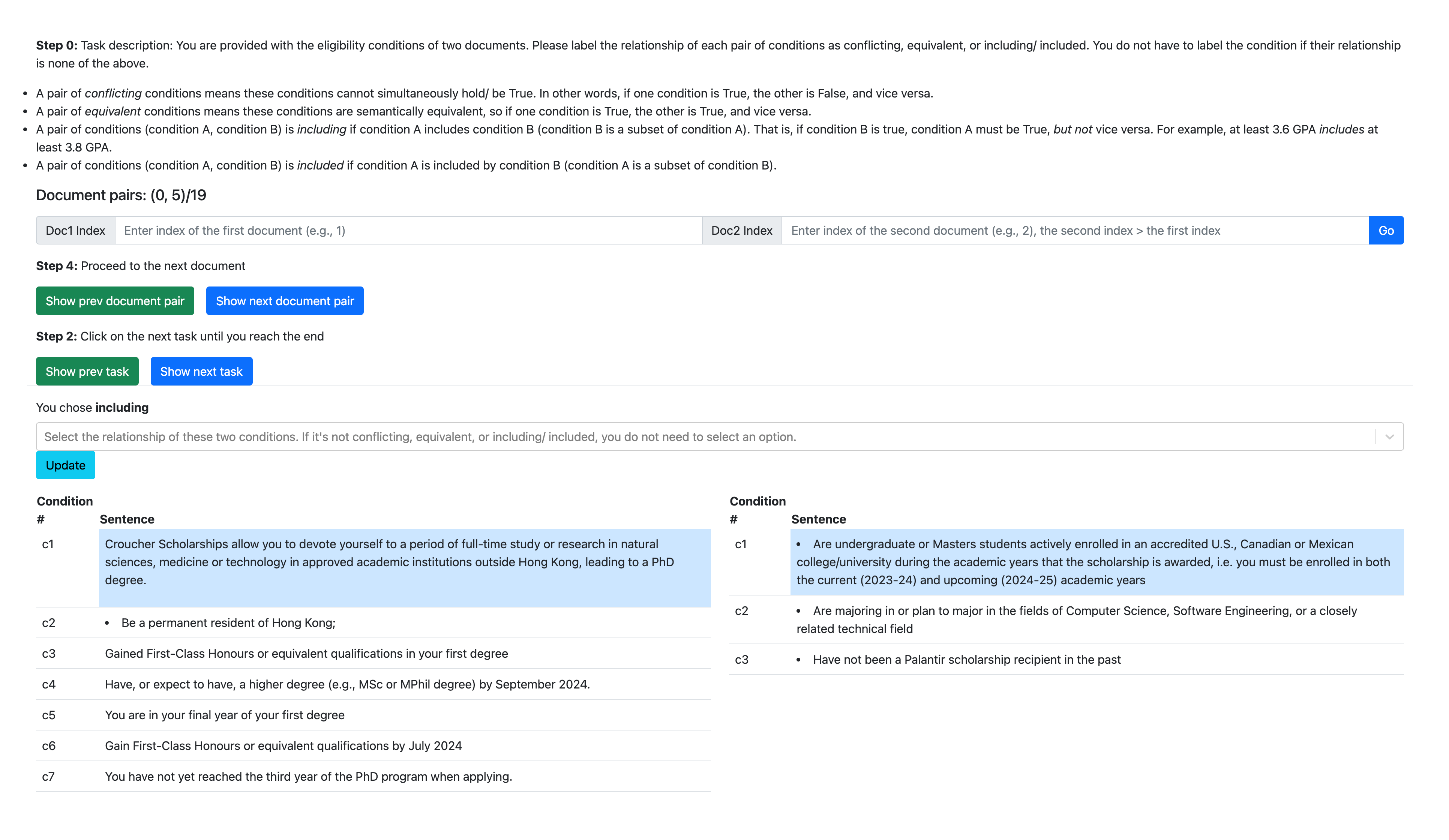}
\caption{Interface for labeling condition relationships.}
\label{fig:annotation_rel}
\end{figure*}

\begin{figure*}
\centering
\includegraphics[width=\textwidth]{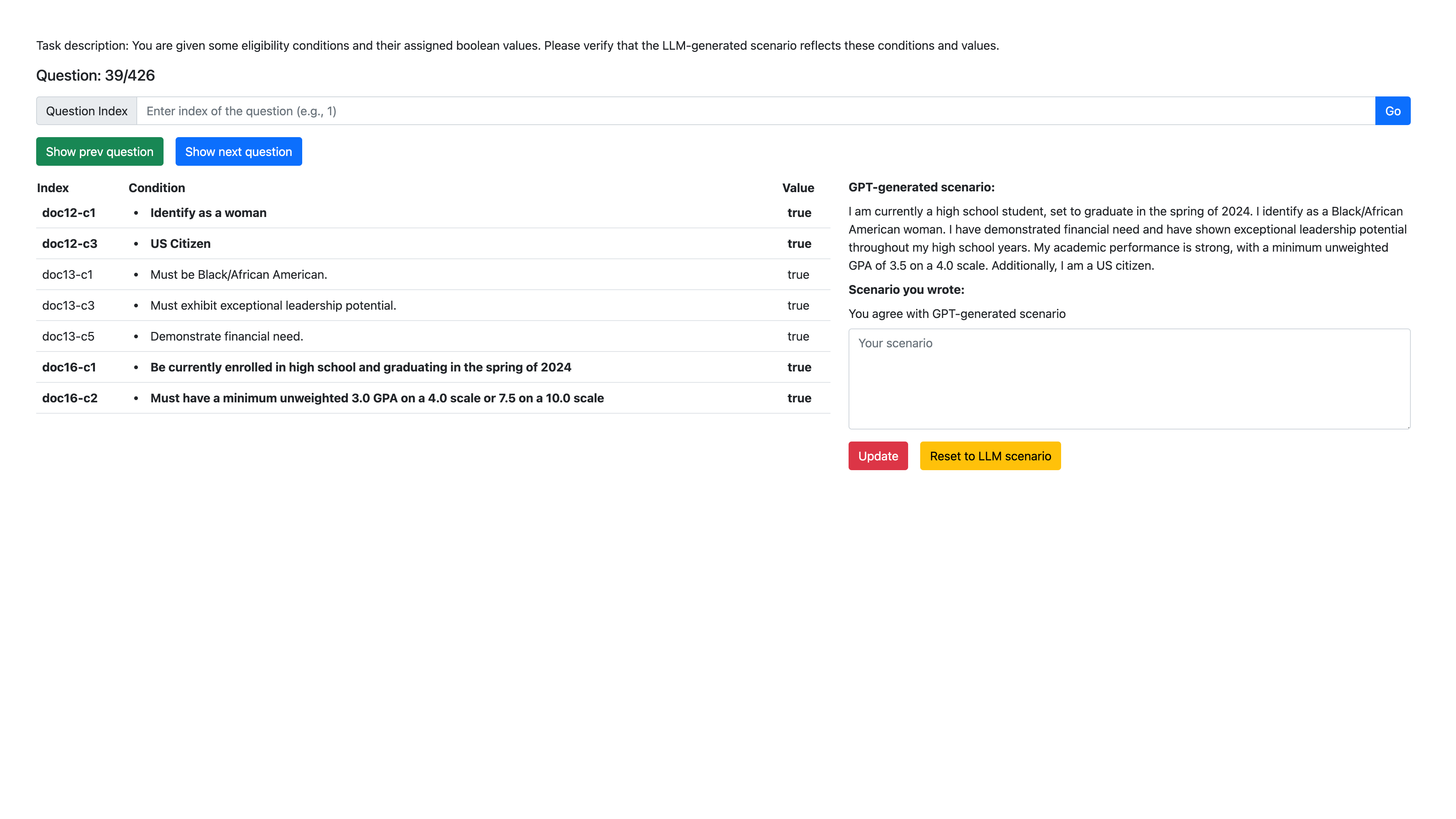}
\caption{Interface for scenario verification.}
\label{fig:annotation_scenario}
\end{figure*}

We show the interfaces annotators use for extracting conditions (Figure \ref{fig:annotation_doc_1}), labeling boolean relationships of relationships within a document (Figure \ref{fig:annotation_doc_2}), labeling condition relationships across documents (Figure \ref{fig:annotation_rel}), and verifying LLM-generated scenarios (Figure \ref{fig:annotation_scenario}).

\section{Performance explanation}
\subsection{Conditional answer analysis}
\label{app:low-f1}

\begin{table*}
\centering
\begin{adjustbox}{max width=\linewidth}
\begin{tabular}{cccccc}
Error types & Relationships  & Predicted (Baseline) & Predicted (With cond relationships) & Correct\\
\midrule
\makecell{Remove conditions \\ across condition groups} & \textbf{doc16-c6} conflicts \textbf{doc19-c0}  & \makecell{[(\textbf{doc16-c6}, doc16-c7, doc16-c9), \\ (\textbf{doc19-c0})]}  & \makecell{[(doc16-c5, doc16-c7, doc16-c8, doc16-c9),  \\ (\textbf{doc19-c0})]}  & \makecell{ [(\textbf{doc16-c6}, doc16-c9, doc16-c10), \\ (\textbf{doc19-c0})]}\\
\midrule
\makecell{Remove conflicting conditions \\ within condition groups} &  \textbf{doc6-c6} conflicts \textbf{doc15-c5} &\makecell{[(\textbf{doc6-c6}, doc6-c7, \\ \textbf{doc15-c5}, doc15-c10)]} & \makecell{[(doc6-c8), \\ (doc15-c7, doc15-c9, doc15-c10)]} & \makecell{[(\textbf{doc6-c6}, doc6-c7), \\ (\textbf{doc15-c5}, doc15-c7, doc15-c10)]}\\
\bottomrule
\end{tabular}
\end{adjustbox}

\caption{Although models understand the given gold information, leveraging provided condition relationships to perform reasoning to obtain correct conditional answers remains challenging.}
\label{tab:rel_drop}
\end{table*}

Table \ref{tab:gold} shows that, generally, providing more gold information as hints in the prompts significantly helps models derive the correct conditional answers. However, providing \circlednum{c} condition relationships does not increase performance. Table \ref{tab:rel_drop} summarizes the error sources.
As described in Section \ref{sec:outputs}, a conditional answer can include multiple groups of unmentioned conditions. A condition group must be \textit{complete} in order to validate the short answer. However, as shown in table \ref{tab:rel_drop}, models tend to remove conditions mentioned in the relationships after explicitly receiving this information and add irrelevant conditions.

First, given condition relationships $r$, models naively remove conditions mentioned in $r$ \textit{across} different condition groups, failing to realize that such removal is unnecessary because the conditions are in different groups and the new condition groups after removal are no longer sufficient to support the short answer. The second case is where models naively remove conditions mentioned in $r$ \textit{within} the condition groups. Models initially fail to recognize the conflicts between benefits, thinking the user is eligible for all, and thus combine unmentioned conditions from \textit{all} documents in the same group. Given the conflict information, the model can realize the conflicts and correctly the single condition group into multiple groups without conflicts. However, models fail to realize that some conditions in $r$ are critical to making new condition groups complete and thus decide not to add the corresponding conditions. This, again, resembles similar issues from case 1, and the new condition groups are still insufficient to support the short answer.

The above behaviors suggest that while models can understand the provided condition relationships and use them in deriving conditional answers to some extent, performing correct reasoning is still a hurdle.

\section{Implementation details}
We used both open-source and commercial language models for benchmarking. The open-source models were loaded from PyTorch and Huggingface, including Llama-3-70B-Instruct \footnote{https://huggingface.co/meta-llama/Meta-Llama-3-70B-Instruct}, Llama-3-8B-Instruct \footnote{https://huggingface.co/meta-llama/Meta-Llama-3-8B-Instruct}, and Gemma-1.1-7B-Instruct \footnote{https://huggingface.co/google/gemma-1.1-7b-it}. Inference using local models was performed on a cluster of 8 V100 GPUs. The commercial model is GPT-4o, and inference using this model was performed using API calls.

\section{Prompts}
\label{app:prompts}

Prompts for scenario generation and benchmarking used in our experiments are included in Table \ref{tab:prompts-scenario}-\ref{tab:prompts-gold}.

\begin{table*}
\centering

\begin{adjustbox}{max width=\linewidth}
\small
\begin{tabular}{p{\linewidth}}
\toprule
You are given several eligibility conditions and values. Your task is to write a factual one-paragraph description from a first-person narrative for a user. The description should just be a summary of conditions that the user satisfies/ does not satisfy according to values given. If a condition can be satisfied several ways, please be concrete and pick only one way to include in your scenario. Do not include information not mentioned in the conditions. Do not mention anything other than these conditions. Do not include "I am eligible to apply" or "I am not eligible to apply". Please also ensure your generated summary is self-consistent and logical.\\
\\
Conditions:\\
1. value: True, condition: Have, or expect to have, a higher degree (e.g., MSc or MPhil degree) by September 2024.\\
2. value: True, condition: Gain First-Class Honours or equivalent qualifications by July 2024\\
3. value: True, condition: <li>Applications must be on behalf of full-time students currently pursuing a PhD at an accredited university in the United States.</li>\\
4. value: True, condition: <li>Students' research must be relevant to one of the disciplines listed below:</li> <li><b>Computer Science</b> - Topics of interest include: machine learning \& artificial intelligence, deep learning, reinforcement learning, natural language processing, computer vision, robotics, computational biology</li>\\
Scenario: I am currently a full-time student pursuing a PhD at an accredited university in the United States. My research is in the field of Computer Science, specifically focusing on machine learning and artificial intelligence. I have achieved First-Class Honours and I have a MSc degree.\\
\\
Conditions:\\
1. value: False, condition: <li>GPA of 3.0 or higher</li>\\
2. value: False, condition: <li>US Citizen</li>\\
3. value: False, condition: Exceptional student members entering the Fall Semester of their Senior Year\\
4. value: False, condition: Citizens of the Republic of the Marshall Islands, Federated States of Micronesia and the Republic of Palau are also eligible to apply\\
5. value: True, condition: <li>Must apply for federal financial aid for the 2024-2025 academic year using the Free Application for the Federal Student Aid (FASFA) by early April 2024</li>\\
Scenario: During preparation for the academic year 2024-2025, I have taken the initiative to apply for federal financial aid by sending my Free Application for the Federal Student Aid (FAFSA) at March 2024. My current standing of GPA is 2.7 and I do not hold the citizenship of the United States. Additionally, I am entering the Fall Semester of my Junior Year and neither do I carry the citizenship of the Republic of the Marshall Islands, Federated States of Micronesia, or the Republic of Palau.\\
\\
Conditions:...\\
Scenario:\\
\bottomrule
\end{tabular}
\end{adjustbox}

\caption{Prompt for scenario generation using GPT-4.}
\label{tab:prompts-scenario}
\end{table*}

\begin{table*}
\centering

\begin{adjustbox}{max width=\linewidth}
\scriptsize
\begin{tabular}{p{\linewidth}}
\toprule
1-shot\\
\midrule
You are provided with information about several documents, including their titles, outcomes, and eligibility conditions of these outcomes. You are also given an user scenario, and an user question. The user scenario may not mention all eligibility conditions stated in the documents. Satisfiability status is undetermined for conditions not explicitly mentioned in the user scenario. Therefore, you should not determine that the user is ineligible for an outcome simply because of unmentioned conditions. Your task is to output a short answer to the user question based on the user scenario and lists of conditions mentioned in the documents but not explicitly mentioned in the user scenario or question that need to be True for your answer to hold. You should output all possible lists of unmentioned conditions that can make the answer hold. If your answer holds regardless of the satisfiability of other unmentioned/ undetermined conditions, you should output your answer and an empty list ([]). You should simplify the list of conditions returned in the output by considering equivalent conditions and inclusive conditions. Two conditions are equivalent if one being True/False implies the other being True/False. Condition A includes condition B if B is a subset of A. If a pair of equivalent conditions need to be True for your answer to hold, you should simplify your answer and only keep one of the two conditions. In the case of inclusive conditions, you should only keep the condition representing the subset. To arrive at the final answer, you should also identify the eligibility conditions from the given documents, check whether these conditions are satisfied based on the user scenario, and consider the effects of conflicting conditions across outcomes. A pair of conflicting conditions means these conditions cannot be simultaneously True; if one condition is True, the other is False, and vice versa. Unsatisfied or conflicting conditions do not necessarily imply that the user is ineligible for all given outcomes. You should output your final answer to the user question after a new line, and the output should consist of both a short answer (it can only be yes or no or a number) and a nested list (e.g., [["doc0-c1", "doc0-c2"], ["doc5-c2"]]) of conditions (please use the syntax of bracket and comma to represent lists). In the nested list of conditions, if all conditions in any sublist are True, then the final answer holds. In other words, each sublist is a conjunction of conditions, and the nested list is a disjunction of the sublists. Each condition should be represented as the concatenation of the doc index and the sentence index wrapped in double quotes (e.g., "doc0-c1"). The answer should be wrapped within the <ans></ans> tags, and the conditions should be wrapped within <cond></cond> tags (e.g., <cond>\textbackslash n[["doc0-c1", "doc0-c2"], ["doc5-c2"]]\textbackslash n</cond>). When you complete, output <FIN></FIN> tags.\\
\\
Here is an example.\\
doc8: APIA SCHOLARSHIP PROGRAM\\
"c0":<h4>ABOUT THE APIA SCHOLARSHIP PROGRAM</h4>\\
"c1":<p>The APIA Scholarship is our largest scholarship program, open to AANHPI undergraduate students attending any U.S. accredited university or college. Scholarship amounts range from $2,500 one-year awards to $20,000 multi-year awards. APIA Scholars has a special focus on supporting AANHPI students who live at or below the poverty line; are in the first generation of their family to attend college; are representative of the APIA community's diversity, (geographically and ethnically), especially those ethnicities that have been underrepresented on college campuses due to limited access and opportunity. Strong applicants would also have an emphasis on community service and leadership.</p>\\
"c2":<h4>APIA SCHOLARS MINIMUM ELIGIBILITY CRITERIA FOR 2024-2025 ACADEMIC YEAR:</h4>\\
"c3":<li>Be able to describe your ethnicity, heritage, or ancestry in relation to the countries, territories, or lands in Asia or the Pacific Islands</li>\\
"c4":<li>Be a citizen, national, or legal permanent resident of the United States. Citizens of the Republic of the Marshall Islands, Federated States of Micronesia and the Republic of Palau are also eligible to apply</li>\\
"c5":<li>Be enrolling or continuing as a degree-seeking undergraduate student in a U.S. accredited college or university in Fall 2024</li>\\
"c6":<li>Have a minimum cumulative GPA of 2.7 on a 4.0 scale (unweighted), or equivalent, or have earned a GED</li>\\
"c7":<li>Must apply for federal financial aid for the 2024-2025 academic year using the Free Application for the Federal Student Aid (FASFA) by early April 2024</li>\\
\\
doc17: Microsoft Women Scholarship\\
"c0":<h2>Scholarship Award</h2>\\
"c1":<p>Seven one-time awards of \$5,000.</p>\\
"c2":<h3><strong>Requirements</strong></h3>\\
"c3":<p>Applicants to the Women at Microsoft scholarship must:</p>\\
"c4":<li>Be a graduating high school senior</li>\\
"c5":<li>Self-identify as a woman*</li>\\
"c6":<li>Plan to enroll in full-time in a tech, engineering, math, or computer science related undergraduate study at an accredited two- or four-year college, university, or vocational-technical school, in the United States, for the entire upcoming academic year.**</li>\\
"c7":<li>Have a minimum grade point average of 3.0 on a 4.0 scale or its equivalent</li>\\
"c8":<p>Employees and children of employees of Microsoft are ineligible.\\
"c9":*Non-binary people, those who are gender fluid, and women of transgender experience are encouraged to apply.\\
"c10":**International applicants are welcome to apply if they will attend school in the US.</p>\\
\\
Scenario: I have earned a GED and have maintained a minimum grade point average of 3.0 on a 4.0 scale. I have not previously received a 9-month fellowship. As a non-binary individual, I am encouraged to apply for this opportunity.\\
Question: What is the maximum number of scholarship(s) I can receive out of the following scholarship(s): APIA SCHOLARSHIP PROGRAM, Microsoft Women Scholarship?\\
Answer:\\
<ans>2</ans>\\
<cond>\\
\lbrack ["doc17-c4", "doc17-c6", "doc17-c8", "doc8-c3", "doc8-c4", "doc8-c7"] \rbrack \\
</cond>\\
<FIN></FIN>\\
\\
\\
Here is the user input.\\
<documents>\\
\\
Scenario: ...\\
Question: What is the maximum number of scholarship(s) I can receive out of the following scholarship(s): ...?\\
Answer:\\
\bottomrule
\end{tabular}
\end{adjustbox}

\caption{1-shot prompt for benchmarking baseline performance.}
\label{tab:prompts-base}
\end{table*}

\begin{table*}
\centering

\begin{adjustbox}{max width=\linewidth}
\scriptsize
\begin{tabular}{p{\linewidth}}
\toprule
0-shot\\
\midrule
You are provided with information about several documents, // \\
\\
Here is the user input.\\
<documents>\\
\\
Scenario: ...\\
Question: What is the maximum number of scholarship(s) I can receive out of the following scholarship(s): ...?\\
Answer:\\
\midrule
1-shot chain-of-thoughts \\
You are provided with information about several documents, // \\
\\
Here is an example.\\
doc8: APIA SCHOLARSHIP PROGRAM // \\
\\
doc17: Microsoft Women Scholarship // \\
\\
Scenario: // \\
Question: // \\
Answer: // \\
\\
\\
Here is the user input.\\
<documents>\\
\\
Scenario: ...\\
Question: What is the maximum number of scholarship(s) I can receive out of the following scholarship(s): ...?\\
\textit{The answer should be wrapped within the <ans></ans> tags, and the conditions should be wrapped within <cond></cond> tags (e.g., <cond>\textbackslash n[["doc0-c1", "doc0-c2"], ["doc5-c2"]]\textbackslash n</cond>). When you complete, output <FIN></FIN> tags. Let's think step by step.} \\
\bottomrule
\end{tabular}
\end{adjustbox}

\caption{0-shot and 1-shot w/ CoT prompt for benchmarking base performance. // means the texts are the same as in Table \ref{tab:prompts-base}.}
\end{table*}

\begin{table*}
\centering
\begin{adjustbox}{max width=\linewidth}
\scriptsize
\begin{tabular}{p{\linewidth}}
\toprule
With \circlednum{a} document conditions\\
\midrule
You are provided with information about several documents, // \\
\\
Here is an example.\\
doc8: APIA SCHOLARSHIP PROGRAM\\
"c3":<li>Be able to describe your ethnicity, heritage, or ancestry in relation to the countries, territories, or lands in Asia or the Pacific Islands</li>\\
"c4":<li>Be a citizen, national, or legal permanent resident of the United States. Citizens of the Republic of the Marshall Islands, Federated States of Micronesia and the Republic of Palau are also eligible to apply</li>\\
"c5":<li>Be enrolling or continuing as a degree-seeking undergraduate student in a U.S. accredited college or university in Fall 2024</li>\\
"c6":<li>Have a minimum cumulative GPA of 2.7 on a 4.0 scale (unweighted), or equivalent, or have earned a GED</li>\\
"c7":<li>Must apply for federal financial aid for the 2024-2025 academic year using the Free Application for the Federal Student Aid (FASFA) by early April 2024</li>\\
\\
doc17: Microsoft Women Scholarship\\
"c4":<li>Be a graduating high school senior</li>\\
"c5":<li>Self-identify as a woman*</li>\\ 
"c6":<li>Plan to enroll in full-time in a tech, engineering, math, or computer science related undergraduate study at an accredited two- or four-year college, university, or vocational-technical school, in the United States, for the entire upcoming academic year.**</li> \\
"c7":<li>Have a minimum grade point average of 3.0 on a 4.0 scale or its equivalent</li> \\
"c8":<p>Employees and children of employees of Microsoft are ineligible.\\
"c9":*Non-binary people, those who are gender fluid, and women of transgender experience are encouraged to apply. \\
\\
Scenario: // \\
Question: // \\
Answer: // \\
\\
\\
Here is the user input.\\
\textit{<documents with relevant conditions only>}\\
\\
Scenario: ...\\
Question: What is the maximum number of scholarship(s) I can receive out of the following scholarship(s): ...?\\
Answer:\\
\midrule
With \circlednum{b} condition satisfiability\\
\midrule
You are provided with information about several documents, including their titles, outcomes, and eligibility conditions of these outcomes. You are also given an user scenario, \textit{satisfiability of some conditions that can be clearly determined based on the user scenario}, and an user question. // \\
\\
Here is an example.\\
doc8: APIA SCHOLARSHIP PROGRAM // \\
\\
doc17: Microsoft Women Scholarship // \\
\\
Scenario: // \\
\textit{Condition satisfiability: doc8-c6 is satisfied. doc17-c7 is satisfied. doc17-c9 is satisfied.}\\
Question: // \\
Answer: // \\
\\
\\
Here is the user input.\\
<documents>\\
\\
Scenario: ...\\
\textit{Condition satisfiability: ...}\\
Question: What is the maximum number of scholarship(s) I can receive out of the following scholarship(s): ...?\\
Answer:\\
\midrule
With \circlednum{c} condition relationships\\
\midrule
You are provided with information about several documents, including their titles, outcomes, and eligibility conditions of these outcomes. You are also given an user scenario, \textit{relationships of conditions across documents}, and an user question. // \\
\\
Here is an example.\\
doc8: APIA SCHOLARSHIP PROGRAM // \\
\\
doc17: Microsoft Women Scholarship // \\
\\
Scenario: // \\
\textit{Condition relationships: doc8-c5 includes doc17-c6.}\\
Question: //\\
Answer: //\\
\\
\\
Here is the user input.\\
<documents>\\
\\
Scenario: ...\\
\textit{Condition relationships: ...}\\
Question: What is the maximum number of scholarship(s) I can receive out of the following scholarship(s): ...?\\
\midrule
With \circlednum{a} + \circlednum{b} + \circlednum{c}\\
\midrule
A combination of the three prompts above.\\
\bottomrule
\end{tabular}
\end{adjustbox}

\caption{Prompts for benchmarking performance when gold information is provided. // means the texts are the same as in Table \ref{tab:prompts-base}.}
\label{tab:prompts-gold}
\end{table*}

\end{document}